\documentclass{article}

\usepackage{microtype}
\usepackage{graphicx}
\usepackage{subcaption}
\usepackage{tabularx}
\usepackage{array}
\usepackage{booktabs} 
\usepackage{makecell} 
\usepackage{multirow}
\usepackage{dblfloatfix}
\usepackage{bbm}

\usepackage{hyperref}

\newcolumntype{L}{>{\raggedright\arraybackslash}X}
\newcolumntype{Y}{>{\centering\arraybackslash}X}

\usepackage[preprint]{icml2026}

\usepackage{amsmath}
\usepackage{amssymb}
\usepackage{mathtools}
\usepackage{amsthm}

\usepackage[capitalize,noabbrev]{cleveref}

\theoremstyle{plain}

\theoremstyle{definition}

\theoremstyle{remark}

\usepackage[textsize=tiny]{todonotes}

\icmltitlerunning{Region-Normalized DPO for Medical Image Segmentation 
  under Noisy Judges}

\begin{document}

\twocolumn[
  \icmltitle{Region-Normalized DPO for Medical Image Segmentation \\
  under Noisy Judges }



  \icmlsetsymbol{equal}{*}

  \begin{icmlauthorlist}
    \icmlauthor{Hamza Kalisch}{ikim,nuk}
    \icmlauthor{Constantin Seibold}{heidel}
    \icmlauthor{Jens Kleesiek}{ikim,tu}
    \icmlauthor{Ken Herrmann}{nuk}
    \icmlauthor{Frederic Jonske}{ikim}
  \end{icmlauthorlist}

  \icmlaffiliation{ikim}{Institute for AI in Medicine (IKIM), University Hospital Essen (AöR), Essen, Germany}
  \icmlaffiliation{nuk}{Department of Nuclear Medicine, University Hospital Essen (AöR), Essen, Germany }
  \icmlaffiliation{tu}{Department of Physics, TU Dortmund, Dortmund, Germany}
  \icmlaffiliation{heidel}{Heidelberg University Hospital, Heidelberg, Germany}

  \icmlcorrespondingauthor{Hamza Kalisch}{hamza.kalisch@uk-essen.de}

  \icmlkeywords{Machine Learning, ICML}

  \vskip 0.3in
]



\printAffiliationsAndNotice{}  

\begin{abstract}
While dense pixel-wise annotations remain the gold standard for medical image segmentation, they are costly to obtain and limit scalability. In contrast, many deployed systems already produce inexpensive automatic quality-control (QC) signals like model agreement, uncertainty measures, or learned mask-quality scores which can be used for further model training without additional ground-truth annotation.
However, these signals can be noisy and biased, making preference-based fine-tuning susceptible to harmful updates. We study Direct Preference Optimization (DPO) for segmentation from such noisy judges using proposals generated by a supervised base segmenter trained on a small labeled set. We find that outcomes depend strongly on how preference pairs are mined: selecting the judge’s top-ranked proposal can improve peak performance when the judge is reliable, but can amplify harmful errors under weaker judges. We propose Region-Normalized DPO (RN-DPO), a segmentation-aware objective which normalizes preference updates by the size of the disagreement region between masks, reducing the leverage of harmful comparisons and improving optimization stability. Across two medical datasets and multiple regimes, RN-DPO improves sustained performance and stabilizes preference-based fine-tuning, outperforming standard DPO and strong baselines without requiring additional pixel annotations.
\end{abstract}

\section{Introduction}
\label{sec:intro}

Medical image segmentation is essential for the development of automated clinical pipelines, supporting tasks such as large scale screening processes \cite{cellina2023artificial}, treatment planning \cite{sherer2021metrics}, and longitudinal monitoring \cite{diaz2022recent}. When trained with dense annotations  segmentation methods achieve strong performance \cite{antonelli2022medical} but for novel tasks such masks are expensive to obtain \cite{wang2021annotation}. To bypass annotation shortcomings alternative annotation settings have been investigated \cite{radsch2024quality}.

A widely used data-collection strategy for NLP settings is preference-based feedback, where users are presented multiple candidate outputs and select the one they prefer most. These pairwise comparisons can then be leveraged to align models, for instance via RLHF \cite{stiennon2020learning,ouyang2022training} or more directly through Direct Preference Optimization (DPO), which provides a simple and effective training objective \cite{rafailov2023direct}.
When translating such a setting to medical imaging, a clinician would compare two candidate masks and indicate which one is better, when producing pixel-level annotation is infeasible.
Initial results \cite{konwer2025enhancing,nam2025model} suggest that comparative supervision via DPO improve medical segmentation without requiring additional dense labels. 

However, despite growing interest in DPO for segmentation, existing studies often rely on preference signals that are effectively derived from ground truth \cite{wu2025sampo, konwer2025enhancing,nam2025model}, which bypassess the challenge of realistic imperfect feedback. In practice, comparative feedback in medical pipelines can arise from quality control (QC) signals, ranging from lightweight human review and heuristic plausibility checks to automated segmentation quality estimation and failure detection models \cite{zenk2025comparative, specktor2025segqc,valindria2017reverse}. Such judges can be instantiated in many forms, including ensemble-agreement based checks \cite{hann2021deep} or consistency signals under test-time perturbations \cite{wang2019aleatoric}.
Beyond evaluation, QC signals have also been used to improve segmentation models, for instance by selecting reliable outputs in ensemble-based pipelines \cite{hann2021deep}, filtering pseudo-labels, or prioritizing samples for annotation in active learning \cite{shi2024predictive}.
Importantly, QC is not perfectly reliable: benchmarking studies show that segmentation QC and failure detection remain challenging and can degrade under distribution shift \cite{zenk2025comparative}, 
while learned QC models exhibit non-negligible error rates in metric estimation and error detection \cite{specktor2025segqc}. This motivates the central question of our work: \textbf{how can we make preference-based fine-tuning for medical image segmentation stable and effective under noisy feedback?}
 
To tackle this question, we study preference-based fine-tuning for medical segmentation under controlled noisy judges. 
Concretely, we instantiate a model-based QC proxy by training an auxiliary segmenter on a separate labeled split. This design yields a realistic source of structured model bias while allowing systematic variation of judge reliability. 
Across two benchmarks, preference fine-tuning is evaluated under different combinations of base strength, judge strength, and pair-mining strategies, revealing that vanilla DPO can be fragile under imperfect rankings and suffer performance degradation from high-leverage misranked comparisons. 
To mitigate this, we introduce Region-Normalized DPO (RN-DPO), which normalizes preference updates by the disagreement region between candidate masks, improving robustness and performance across regimes. 
Our contributions are: (i) a systematic empirical study of DPO for medical image segmentation under noisy comparative feedback, (ii) RN-DPO, a segmentation-aware objective that stabilizes preference fine-tuning under judge noise, and (iii) consistent improvements over vanilla DPO and other strong baselines across different judge strengths, including strong gains under an oracle judge.
\section{Related Work}
\label{sec:formatting}


\paragraph{Direct Preference Optimization and its variants.}
Direct Preference Optimization (DPO) \cite{rafailov2023direct} casts preference learning as a simple binary classification objective
derived from a KL-regularized reward-maximization view, directly optimizing a policy against a fixed reference model and
thereby avoiding explicit reward modeling and online RL.
A number of extensions refine DPO-style training via alternative parameterizations \cite{tang2024generalized}, offsets \cite{amini2024direct}, or reference-free
formulations \cite{meng2024simpo, hong2024orpo}. Token-level variants such as TDPO \cite{zeng2024token} further increase
granularity by applying preference constraints within sequences, and closely related objectives like IPO
\cite{azar2024general} emphasize stability through alternative regularization choices. 

\paragraph{Preference optimization for vision and segmentation.}
Preference optimization has also been adapted to vision-related tasks, for example to align text-to-image diffusion models \cite{wallace2024diffusion} and to improve vision-language model faithfulness \cite{xie2024v}. In medical image segmentation, \cite{konwer2025enhancing} apply preference optimization to further fine-tune a SAM-based model \cite{kirillov2023segment} using comparative supervision over candidate masks generated by thresholding the output probabilities of the encoder's prediction. Similarly, SAMPO \cite{wu2025sampo} applies DPO to promptable vision foundation models by mining preference pairs from candidate masks generated under varied sparse prompts. MAPO \cite{nam2025model} extends this to arbitrary segmentation backbones via a model-agnostic framework and utilizes dropout to sample more diverse candidates. While these works demonstrate the feasibility of learning from comparative signals for the segmentation task, they rely on ground-truth masks to define preferences. With only limited rating-noise ablations reported in \cite{konwer2025enhancing}, robustness under noisy mined preferences in segmentation remains largely unexplored.


\paragraph{Robustness under noisy preferences.}
Preference data can be noisy in various ways (e.g., flipped comparisons, near-ties, judge-specific biases), which can lead to performance degradation and instability \cite{chowdhury2024provably}. rDPO \cite{chowdhury2024provably} provides a theoretically grounded robust loss for DPO under random label flips, while Dr.\ DPO \cite{wu2024towards} adopts a distributionally robust view to handle unreliable pairs. Complementary approaches improve robustness via adaptive reweighting and calibration, e.g., $\beta$-DPO \cite{wu2024beta},
which dynamically adjusts the DPO temperature to reduce sensitivity to noisy comparisons.
In contrast, $\alpha$-DPO \cite{wu2024alpha} and $\gamma$-PO \cite{sun2025robust} introduce margin-based formulations, using adaptive or pair-specific target margins to down-weight ambiguous preference pairs. Beyond objective design, robustness can be influenced by how preference pairs are selected or scheduled, e.g. via a curriculum sampling of easier vs harder pairs \cite{croitoru2025curriculum}.

\begin{figure*}[t]
    \centering
    \includegraphics[width=\textwidth, 
  trim=0.5cm 1.5cm 1.25cm 0cm,
  clip]{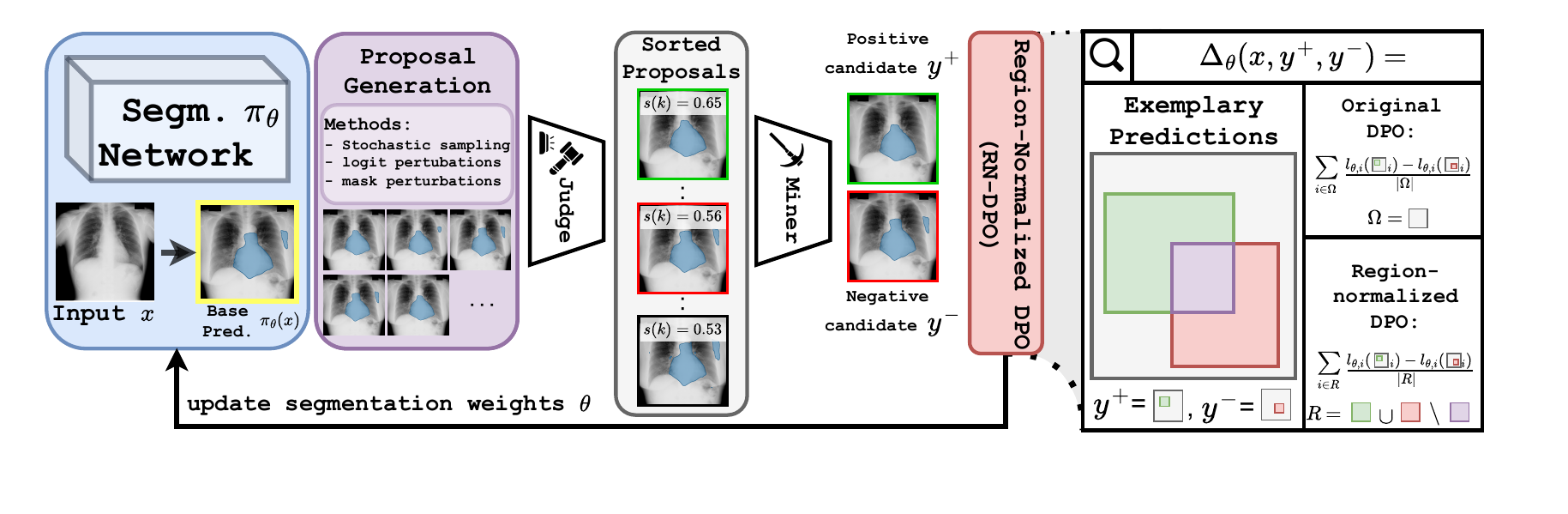}
    \caption{\textbf{Main pipeline.} We fine-tune a base segmenter $\pi_{\theta}$ on unlabeled images using comparative feedback. For each image, $K$ candidate masks are generated, scored by a judge, and converted into preference pairs by a miner. The base segmenter is then updated with RN-DPO, which normalizes the preference update over the disagreement region between the two masks instead of the whole image. Notation follows Sec. \ref{sec:method}.}
    \label{fig:main}
\end{figure*}

\paragraph{Quality control and segmentation quality estimation.}
Medical segmentation pipelines often incorporate QC mechanisms or quality-estimation models \cite{robinson2019automated} that can
provide cheap scalar scores or rankings without utilizing dense labels. 
RCA \cite{valindria2017reverse} and its in-context variant \cite{cosarinsky2025context} estimate segmentation quality using a labeled reference set as an indirect proxy, while learned QC predictors directly score segmentation quality and can highlight likely error regions \cite{fournel2021medical, specktor2025segqc}.  Complementary “label-free” QC signals such as confidence aggregation
and ensemble disagreement have also been shown to be effective failure detectors under distribution shift
\cite{zenk2025comparative}. QC estimates have also been used for pseudo-label filtering or to prioritize samples for annotation in active learning \cite{ruijsink2020quality, shi2024predictive},
whereas we utilize QC as a comparative signal to rank candidate masks for preference optimization.

\section{Method}
\label{sec:method}

Our approach consists of two stages. In the first stage, a base segmenter is trained on a small labeled dataset in a supervised fashion. In the second stage (see Fig. \ref{fig:main}), the segmenter is improved using comparative feedback without utilizing additional pixel-wise labels. For each unlabeled image, the current segmenter generates a slate of diverse candidate masks and a judge assigns scores to these candidates to induce an ordering over the slate. A miner selects preference pairs from the scored slate, and the segmenter is fine-tuned using a DPO-style objective. 

\subsection{Preliminaries}
\label{sec:prelim}

\textbf{Data, budgets, and notation.}
Let $\mathcal{D}_{\text{seg}}=\{(x_i,y_i)\}_{i=1}^{N_{\text{seg}}}$ denote the pixel-labeled set used to train a base segmenter, where $x$ is an image and $y$ is a segmentation mask.
Let $\mathcal{D}_{\text{pref}}=\{x_j\}_{j=1}^{N_{\text{pref}}}$ denote unlabeled images used in the second stage for preference fine-tuning.
To instantiate controlled noisy comparative feedback, an auxiliary judge is trained on a separate pixel-labeled split of size $N_{\text{QC}}$.
These labels are used solely to train the judge and are never used as pixel-wise supervision for the segmenter.
During preference fine-tuning, the segmenter is optimized solely from preference pairs derived from the judge scores. 

\textbf{Segmenter and per-pixel likelihood.}
The segmenter $\pi_\theta$ maps an image $x$ to logits $z_\theta(x)\in\mathbb{R}^{C\times H\times W}$.
Let $\Omega$ denote the set of spatial indices (pixels/voxels).

\emph{Multi-class.}
Let $\pi_{\theta,i}=\mathrm{softmax}(z_\theta(x)_{:,i})$ be the categorical distribution at index $i\in\Omega$.
Then, for a candidate mask $y\in\{1,\dots,C\}^{H\times W}$, the log-likelihood is given by 
\begin{equation}
\log \pi_{\theta}(y\mid x)
=\frac{1}{|\Omega|}\sum_{i\in\Omega}\log \pi_{\theta,i}(y_i).
\label{eq:seg_ll_mc}
\end{equation}

\emph{Multi-label.}
For $C$ independent binary labels, let $p_{\theta,i,c}=\sigma(z_\theta(x)_{c,i})$.
For a fixed label $c$, we define the per-pixel averaged Bernoulli log-likelihood
\begin{equation}
\log \pi_{\theta}(y_c\mid x)
=\frac{1}{|\Omega|}\sum_{i\in\Omega}\log \mathrm{Bern}\!\left(y_{i,c};\, p_{\theta,i,c}\right).
\label{eq:seg_ll_ml}
\end{equation}

Let $\ell_{\theta,i}(\cdot)$ denote the per-pixel log-likelihood term under $\pi_\theta$:
$\ell_{\theta,i}(y_i)=\log \pi_{\theta,i}(y_i)$ for multi-class, and
$\ell_{\theta,i}(y_{i,c})=\log \mathrm{Bern}(y_{i,c};\,p_{\theta,i,c})$ for a fixed label $c$.

\subsection{Preference Fine-Tuning from Noisy Judges}
\label{sec:rn_dpo}

\textbf{Proposal Generator.}
Given an input image $x$, we generate a slate of $K$ candidate masks $\{y^{(k)}\}_{k=1}^K$ from the current segmenter and lightweight perturbations designed to provide headroom (i.e., candidates that can improve over the current prediction). Besides the base prediction, the slate combines a mixture of various proposal types, including:
\vspace{-1mm}
\begin{itemize}\setlength{\itemsep}{0pt}\setlength{\parskip}{0pt}\setlength{\topsep}{1pt}
\item stochastic model proposals (e.g., dropout / TTA )
\item logit-space perturbations (e.g., temperature shifts)
\item probability-space thresholding
\item simple morphology/topology edits \\(connected-component removal, hole filling)
\item boundary perturbations via signed distance fields
\end{itemize}
\vspace{-1mm}
For a more detailed overview, we refer to the Appendix \ref{app:proposals}. Unless stated otherwise, slates are generated online from the current model during preference fine-tuning, inducing a moving proposal distribution over training.

\textbf{Noisy judges and slate ranking.}
Given an image $x$ and a slate of candidate masks, the judge produces its own segmentation prediction
$\hat{y}_J(x)$ using a separate segmentation model trained on a QC-labeled split of size $N_{\text{QC}}$.
Each proposal is assigned a scalar score based on its agreement with the judge prediction,
$s^{(k)} = J(x, y^{(k)}) := \mathrm{IoU}\!\left(y^{(k)}, \hat{y}_J(x)\right)$,
which induces a ranking over the slate.
We instantiate \emph{weak} and \emph{strong} judges by varying the budget $N_{\text{QC}}$ for the judge.
Importantly, judge supervision is never used to train the segmenter; it is only used to generate noisy preference rankings over proposals.

\textbf{Pair mining.}
Given a ranked slate, a miner $M$ produces a set of preference pairs
$\mathcal{P}(x) = \{(y^+,y^-)\}$, where $y^+$ is preferred over $y^-$ according to the judge.
Miners trade off coverage versus risk under noise. We consider: 
\vspace{-1mm}
\begin{itemize}\setlength{\itemsep}{0pt}\setlength{\parskip}{0pt}\setlength{\topsep}{1pt}
\item \emph{Top-vs-Base:} the top-ranked proposal is paired with the model's current/base prediction.
\item \emph{Top-vs-Random:} the top-ranked proposal is paired with a randomly sampled lower-ranked candidate to provide diverse negatives.
\item \emph{Threshold mining:} the top-ranked proposal is paired with the first lower-ranked candidate whose judge score is at least a fixed margin $\tau$ below the top-ranked score.
\item \emph{Random mining:} samples a random proposal as $y^+$ and a second random proposal ranked below it as $y^-$. 
\end{itemize}
\vspace{-1mm}

\textbf{Standard DPO for segmentation.}
The segmenter can be optimized using these mined preferences via DPO.
Let $\pi_{\theta}$ denote the current segmenter and $\pi_{\mathrm{ref}}$ a fixed reference segmenter (typically the Stage~1 checkpoint).
For each mined preference pair $(y^+,y^-)$ for image $x$, DPO optimizes a logistic objective on the relative log-likelihood ratio:
\begin{align}
\Delta_{\theta}(x;y^+,y^-) &= \log \pi_{\theta}(y^+\mid x) - \log \pi_{\theta}(y^-\mid x), \\
\Delta_{\mathrm{ref}}(x;y^+,y^-) &= \log \pi_{\mathrm{ref}}(y^+\mid x) - \log \pi_{\mathrm{ref}}(y^-\mid x), \\
\mathcal{L}_{\mathrm{DPO}} &= -\mathbb{E}_{(x,y^+,y^-)}\Big[\log \sigma\big(\beta(\Delta_{\theta}-\Delta_{\mathrm{ref}})\big)\Big],
\label{eq:dpo_loss}
\end{align}
where $\sigma(\cdot)$ is the sigmoid and $\beta>0$ controls the strength of the preference update.

\subsection{Region-Normalized DPO}
\label{sec:rn_dpo}

The per-pixel log-ratio is defined by
\begin{equation}
\delta_{\theta,i}\;:=\;\ell_{\theta,i}(y_i^+)-\ell_{\theta,i}(y_i^-).
\end{equation}
Pixels where $y_i^+=y_i^-$ contribute the same term to both $\log\pi_\theta(y^+\mid x)$ and $\log\pi_\theta(y^-\mid x)$ and therefore cancel in the difference, yielding
\begin{equation}
\Delta_\theta(x;y^+,y^-)
=\frac{1}{|\Omega|}\sum_{i\in\Omega}\delta_{\theta,i}
=\frac{1}{|\Omega|}\sum_{i\in R}\delta_{\theta,i},
\label{eq:delta_support_R}
\end{equation}
where the disagreement region (computed from $(y^+,y^-)$) is
\begin{equation}
R\;=\;\{\,i\in\Omega:\; y_i^+\neq y_i^-\,\}.
\label{eq:def_R}
\end{equation}

We assume $|R|>0$. For multi-label datasets, we apply all definitions below independently to each class $c$ using binary masks
$(y_c^+,y_c^-)$ and disagreement regions $R_c$, and average the loss over mined class-level preference pairs.
For readability, we omit the class index $c$.

\textbf{Region-normalized likelihood.}
While vanilla DPO uses the global normalization $1/|\Omega|$, RN-DPO normalizes over the disagreement region:
\begin{equation}
\log \pi_\theta^{R}(y\mid x)
=\frac{1}{|R|}\sum_{i\in R}\ell_{\theta,i}(y_i).
\label{eq:ll_R}
\end{equation}

\textbf{RN-DPO objective.}
Analogous to vanilla DPO, we define the region-normalized log-likelihood ratios for the current and reference models as
\begin{equation}
\begin{aligned}
\Delta_{\theta}^{R}
&:= \log \pi_{\theta}^{R}(y^{+}\!\mid x)\;-\;\log \pi_{\theta}^{R}(y^{-}\!\mid x),\\
\Delta_{\mathrm{ref}}^{R}
&:= \log \pi_{\mathrm{ref}}^{R}(y^{+}\!\mid x)\;-\;\log \pi_{\mathrm{ref}}^{R}(y^{-}\!\mid x).
\end{aligned}
\label{eq:delta_R_defs}
\end{equation}
RN-DPO is obtained by substituting $\Delta^{R}_{\theta}$ and $\Delta^{R}_{\mathrm{ref}}$ into \eqref{eq:dpo_loss}:
\begin{equation}
\mathcal{L}_{\mathrm{RN\mbox{-}DPO}}
= -\mathbb{E}\!\left[\log \sigma\!\left(\beta(\Delta^{R}_{\theta}-\Delta^{R}_{\mathrm{ref}})\right)\right].
\label{eq:rn_dpo}
\end{equation}
With our definitions (multi-class, or multi-label
), the global- and region-normalized ratios are related by
\begin{equation}
\Delta_\theta(x;y^+,y^-)=\frac{|R|}{|\Omega|}\,\Delta_\theta^{R}(x;y^+,y^-),
\label{eq:delta_scale}
\end{equation}
so vanilla DPO implicitly scales each pair by the disagreement fraction $|R|/|\Omega|$, which RN-DPO removes.
Intuitively, RN-DPO decouples update magnitude from disagreement area: large-disagreement (and potentially misranked) comparisons are prevented from dominating updates, while preserving learning signal when candidates are close.

\section{Experimental Setup}

\subsection{Datasets and evaluation metrics}\label{subsec:datasets}
We evaluate on two established medical segmentation benchmarks covering both multi-label and multi-class settings. 
Across the datasets, we construct weak/strong judge regimes via a disjoint QC-labeled pool of 10 cases (images/subjects) for judge training and varying the judge budget by sampling $N_{\text{QC}}$ examples from this pool; these labels are used solely for judge training and are never used as pixel-wise supervision for the segmenter during preference fine-tuning.

\textbf{JSRT.}
We use the JSRT \cite{shiraishi2000development} chest radiograph dataset for multi-label segmentation of thoracic structures, including the heart, lungs (left/right), and clavicles (left/right). All images are resized to $512\times512$ and intensity-normalized using min-max scaling. We split the data into 157 training images, 30 validation images (used for model selection and hyperparameter tuning), and 50 test images (held out for final evaluation). From the 157 training images, we randomly sample $N_{\text{seg}}$ images (with pixel labels) to form $\mathcal{D}_{\text{seg}}$ for supervised base training, and use the remaining training images as the unlabeled pool $\mathcal{D}_{\text{pref}}$ for preference fine-tuning.

\textbf{ACDC.}
We use the ACDC \cite{bernard2018deep} cardiac MRI dataset for multi-class segmentation. Following a 2D training protocol, we treat each axial slice as a training sample and use a 2D U-Net backbone for all methods for a controlled comparison. All images are resized to $256\times256$ and intensity-normalized using min-max scaling. The split comprises 60 training-, 10 validation- (for model selection and hyperparameter tuning), and 20 test-subjects (held out for final evaluation). From the training split, we randomly sample $N_{\text{seg}}$ labeled subjects to form $\mathcal{D}_{\text{seg}}$ for supervised base training, and use the remaining training subjects as the unlabeled pool $\mathcal{D}_{\text{pref}}$ for preference fine-tuning. For ACDC, $N_{\text{QC}}$ counts slices from the QC-labeled pool.

\textbf{Evaluation metrics.}
We report peak segmentation accuracy to measure performance, and a tail-average metric to capture robustness to late-epoch performance degradation.
For peak \textbf{IoU}, we select the checkpoint with highest validation IoU and report its corresponding test IoU.
For \textbf{TailAvg}, we smooth the validation IoU curve with a moving average with a window size of three and average the final six epochs.

\subsection{Implementation Details}\label{subsec:impl_details}

\textbf{Stage 1 (base training).} We train a 2D U-Net with supervised pixel-wise losses on $N_{\text{seg}}$ labeled examples. We use random resized crops, small rotations, and mild photometric jitter (brightness/contrast and gamma) for JSRT, while only applying random flips and rotations for ACDC. Optimization uses AdamW with learning rate $1\times 10^{-3}$, weight decay $1\times 10^{-4}$, and a cosine learning-rate schedule with warmup. The checkpoint with the best validation IoU defines the frozen reference model $\pi_{\theta_0}$. 

\textbf{Stage 2 (preference fine-tuning).} We keep $\pi_{\theta_0}$ fixed and fine-tune $\pi_\theta$ using mined preference pairs. We optimize with Adam, a constant learning rate, batch size of 8, and apply no data augmentations. We tune the learning rate and $\beta$ per dataset, judge regime, and method, as we observed that DPO and RN-DPO can prefer different learning-rate scales. The full hyperparameter grid and selected values are reported in the Appendix \ref{app:hpms}. All other optimization and mining settings are held fixed. We exclude uninformative slates (e.g., zero-valued or near-tied judge scores) by skipping pairs with judge score difference between of $10^{-4}$, to avoid training on comparisons without signal.

\subsection{Regimes}
We evaluate four regimes formed by crossing base and judge strength: \{weak, strong\}$\times$\{weak, strong\}. Base strength is set by the Stage~1 labeled budget $N_{\text{seg}}$. Judge strength is set by the QC budget $N_{\text{QC}}$, with weak/strong judges defined separately for weak vs.\ strong bases. This is necessary due to stronger bases generate higher-quality, more similar proposals, so a fixed absolute $N_{\text{QC}}$ would induce different effective ranking noise across base strengths. We, thus, choose $N_{\text{QC}}$ per base so that the judge is weak or strong \emph{relative to the corresponding proposal distribution}, and report per-regime judge reliability diagnostics in the Appendix \ref{app:diagnostics}.

\subsection{Baselines and methods}

\textbf{Select-best} outputs the highest-scoring \emph{proposal} under the judge without updating the segmenter.
\textbf{Top-1 Pseudolabeling} fine-tunes the segmenter on the highest-scoring proposal as a hard target using the same supervised loss as in Stage~1.
We evaluate standard \textbf{DPO}~\cite{rafailov2023direct} with our default Top-vs-Random mining protocol and a Random mining variant.
We further evaluate \textbf{IPO}~\cite{azar2024general} as an alternative preference objective to DPO and \textbf{rDPO}~\cite{chowdhury2024provably}, which accounts for noisy preference flips.
Our proposed \textbf{RN-DPO} is evaluated using the same mining protocols (Top-vs-Random and Random).
\setlength{\abovecaptionskip}{0pt}
\setlength{\belowcaptionskip}{0pt}

\begin{table}[!b]
\centering
\caption{\textbf{Miner ablation on JSRT (weak base, weak judge)}. Entries are means over 3 label-draw seeds.}
\label{tab:miner_ablation_jsrt_weak}
\setlength{\tabcolsep}{6pt}
\renewcommand{\arraystretch}{1.08}
\small
\begin{tabular}{lcc}
\toprule
\textbf{Miner} & \textbf{IoU} $\uparrow$ & \textbf{TailAvg} $\uparrow$ \\
\midrule
Supervised        & 0.558$\pm$0.017 & / \\
\midrule
Top-vs-Base        & 0.599$\pm$0.006 & 0.503$\pm$0.041  \\
Top-vs-Random      & 0.604$\pm$0.002 & 0.565$\pm$0.011 \\
Threshold mining   & 0.604$\pm$0.010 & 0.582$\pm$0.014  \\
Random mining      & 0.608$\pm$0.019 & 0.594$\pm$0.015 \\
\bottomrule
\end{tabular}
\end{table} 
\begin{table*}[!b]
\centering
\caption{\textbf{JSRT main results} (mean$\pm$std over 3 label-draw seeds).
IoU and TailAvg are reported for various methods across four different regimes, that are characterized by weak/strong base segmenters and judges. $N_{\text{QC}}$ are chosen separately for weak and strong bases to maintain comparable judge reliability across regimes.}
\label{tab:jsrt_main}
\setlength{\tabcolsep}{5pt}
\renewcommand{\arraystretch}{1.12}
\small


\begin{tabularx}{\textwidth}{Y l c cc cc}
\toprule
\multirow{2}{*}{\textbf{Regime}} & \multirow{2}{*}{\textbf{Method}} &
\multirow{2}{*}{\makecell{\textbf{$\boldsymbol{N_{\text{QC}}}$}\\[-1pt] (weak/strong)}} &
\multicolumn{2}{c}{\textbf{Weak Base} ($N_{\text{seg}}{=}2$)} &
\multicolumn{2}{c}{\textbf{Strong Base} ($N_{\text{seg}}{=}6$)} \\
\cmidrule(lr){4-5}\cmidrule(lr){6-7}
& & &
\textbf{IoU} $\uparrow$ & \textbf{TailAvg} $\uparrow$ &
\textbf{IoU} $\uparrow$ & \textbf{TailAvg} $\uparrow$ \\
\midrule

\multicolumn{2}{l}{\textbf{Supervised baseline}} &
--  &
0.558$\pm$0.017 &  --&
0.730$\pm$0.013 &  -- \\
\midrule
\multirow{7}{*}{\textbf{Weak judge}}
& Select-best             & \multirow{7}{*}{\makecell{3/6}} &
0.602$\pm$0.011 & -- &
0.741$\pm$0.015 & -- \\
& Top-1 Pseudolabeling        & &
0.623$\pm$0.010 & 0.607$\pm$0.016 &
0.747$\pm$0.002 & 0.740$\pm$0.003 \\
& DPO \cite{rafailov2023direct}  & &
0.604$\pm$0.002 & 0.565$\pm$0.011 &
0.745$\pm$0.007 & 0.729$\pm$0.017 \\
& DPO (rand.)             & &
0.608$\pm$0.019 & 0.594$\pm$0.015 &
0.757$\pm$0.007 & 0.730$\pm$0.008 \\
& IPO \cite{azar2024general}      & &
0.606$\pm$0.004 & 0.588$\pm$0.004 &
0.750$\pm$0.003 & 0.726$\pm$0.014 \\
& rDPO \cite{chowdhury2024provably} & &
0.602$\pm$0.002 & 0.562$\pm$0.005 &
0.746$\pm$0.003 & 0.720$\pm$0.013 \\
& RN-DPO                  & &
0.631$\pm$0.012 & 0.619$\pm$0.009 &
0.762$\pm$0.007 & 0.759$\pm$0.010 \\
& RN-DPO (rand.)          & &
\textbf{0.634$\pm$0.012} & \textbf{0.620$\pm$0.013} &
\textbf{0.764$\pm$0.007} & \textbf{0.767$\pm$0.015} \\
\midrule

\multirow{7}{*}{\textbf{Strong judge}}
& Select-best            & \multirow{7}{*}{\makecell{6/9}} &
0.623$\pm$0.011 & -- &
0.759$\pm$0.014 & -- \\
& Top-1 Pseudolabeling             & &
\textbf{0.732$\pm$0.008}  & \textbf{0.713$\pm$0.001} &
0.790$\pm$0.007  & 0.781$\pm$0.004 \\
& DPO \cite{rafailov2023direct}                     & &
0.696$\pm$0.007 & 0.679$\pm$0.007 &
0.766$\pm$0.009 & 0.743$\pm$0.009 \\
& DPO (rand.)             & &
0.673$\pm$0.007 & 0.642$\pm$0.003 &
0.762$\pm$0.002 & 0.717$\pm$0.002 \\
& IPO \cite{azar2024general}      & &
0.701$\pm$0.007 & 0.685$\pm$0.010 &
0.768$\pm$0.003 & 0.751$\pm$0.011 \\
& rDPO \cite{chowdhury2024provably}                    & &
0.697$\pm$0.008 & 0.677$\pm$0.008 &
0.755$\pm$0.005 & 0.739$\pm$0.012 \\
& RN-DPO                  & &
0.721$\pm$0.005 & 0.704$\pm$0.003 &
\textbf{0.798$\pm$0.007} & \textbf{0.789$\pm$0.005} \\
& RN-DPO (rand.)          & &
0.717$\pm$0.004 & 0.694$\pm$0.003 &
0.791$\pm$0.016 & 0.781$\pm$0.007 \\
\midrule

\multirow{2}{*}{\textbf{GT oracle}}
& DPO                     & -- &
0.771$\pm$0.001 & 0.758$\pm$0.005 &
0.800$\pm$0.013 & 0.796$\pm$0.017 \\
& RN-DPO                  & -- &
\textbf{0.820$\pm$0.011} & \textbf{0.816$\pm$0.011} &
\textbf{0.850$\pm$0.004} & \textbf{0.855$\pm$0.004} \\
\bottomrule
\end{tabularx}
\end{table*} 

\section{Results}\label{sec:results}

\subsection{Mining strategies}\label{subsec:miners}
We first study how pair mining affects preference fine-tuning performance and stability under noisy judges. Table~\ref{tab:miner_ablation_jsrt_weak} shows that mining strategy can substantially influence late-stage validation behavior of vanilla DPO in the weak-judge regime. Interestingly, even simple random pair selection can perform competitively in this setting, suggesting that broader pair coverage may reduce reliance on potentially misranked top selections. At the same time, more selective miners (e.g., thresholding) can improve stability by filtering ambiguous comparisons, while aggressive top-ranked mining can still exhibit pronounced degradation when the judge occasionally mis-ranks candidates, consistent with rare but high-leverage harmful updates. Overall, these results indicate that mining choices can mitigate risk but do not fully resolve brittleness under noisy comparative feedback, motivating a more robust objective that reduces the impact of such failures (Sec.~\ref{subsec:main}). Unless stated otherwise, we use the Top-vs-Random miner as the default miner in all subsequent experiments, since selecting the top-ranked candidate is the most natural choice to maximize headroom under a reliable judge, while random lower-ranked negatives retain broader pair coverage and reduce sensitivity to occasional top-selection errors.
\setlength{\abovecaptionskip}{0pt}
\setlength{\belowcaptionskip}{-10pt}

\begin{figure}[t]
\centering

\begin{subfigure}[t]{\linewidth}
  \centering
  \begin{minipage}[t]{0.49\linewidth}
    \centering
    \includegraphics[width=\linewidth]{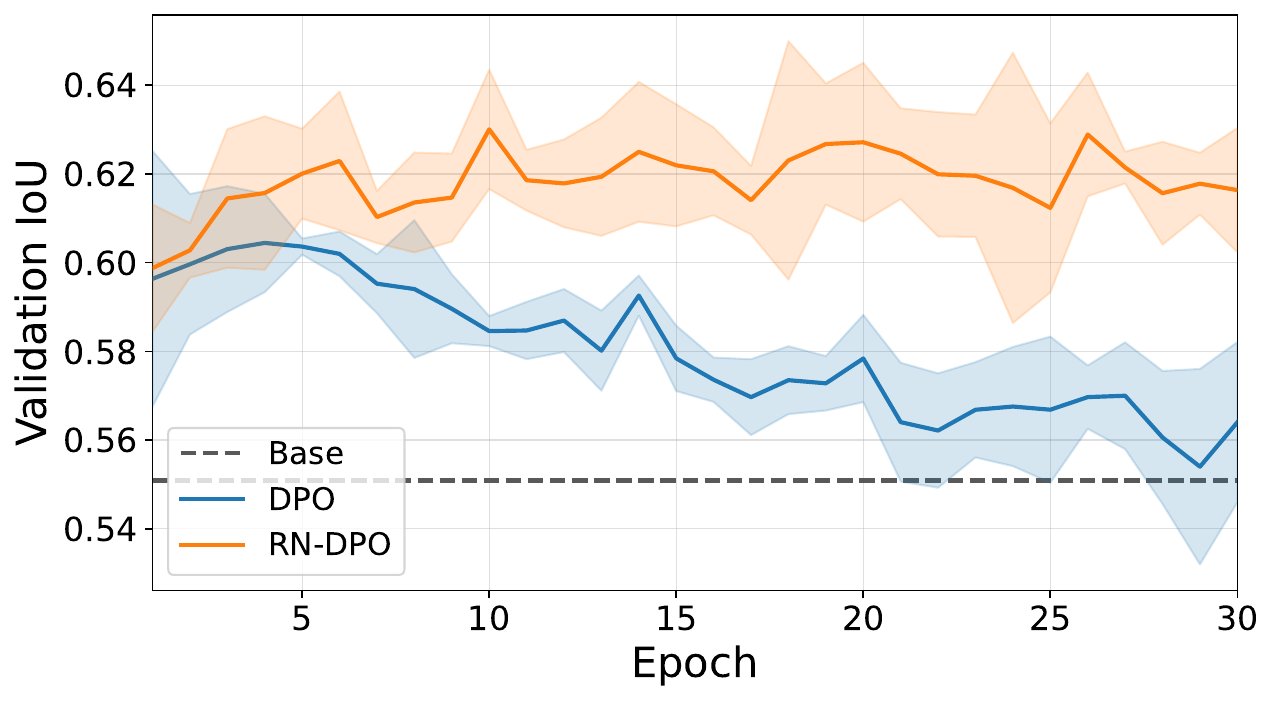}
    \subcaption*{Weak judge}
  \end{minipage}\hfill
  \begin{minipage}[t]{0.49\linewidth}
    \centering
    \includegraphics[width=\linewidth]{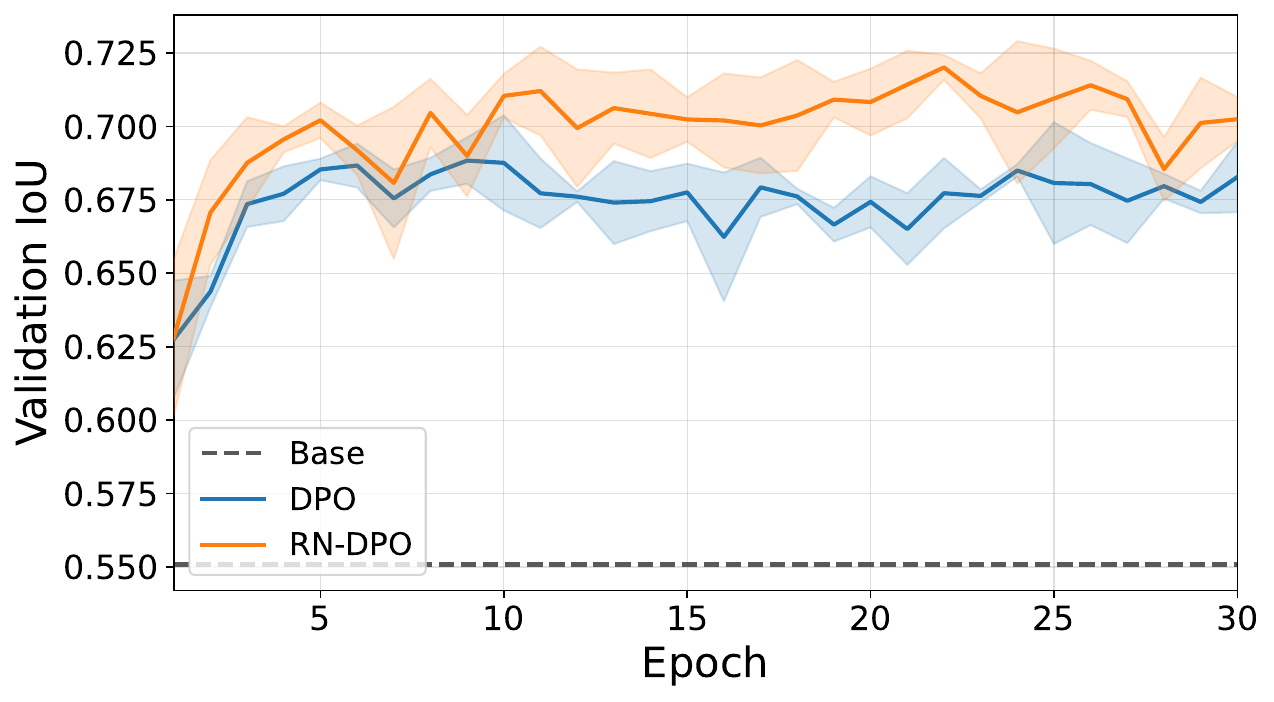}
    \subcaption*{Strong judge}
  \end{minipage}

  \caption{\textbf{JSRT}}
  \label{fig:curves_jsrt}
\end{subfigure}

\vspace{1em}

\begin{subfigure}[t]{\linewidth}
  \centering
  \begin{minipage}[t]{0.49\linewidth}
    \centering
    \includegraphics[width=\linewidth]{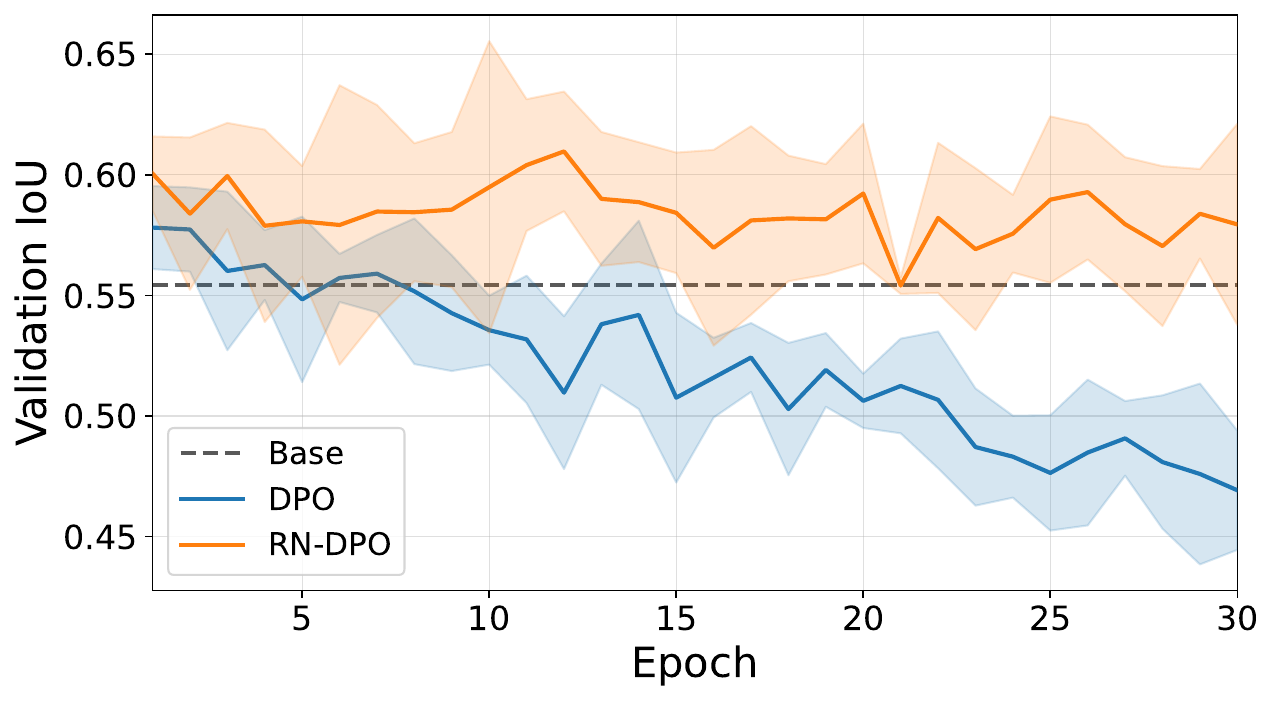}
    \subcaption*{Weak judge}
  \end{minipage}\hfill
  \begin{minipage}[t]{0.49\linewidth}
    \centering
    \includegraphics[width=\linewidth]{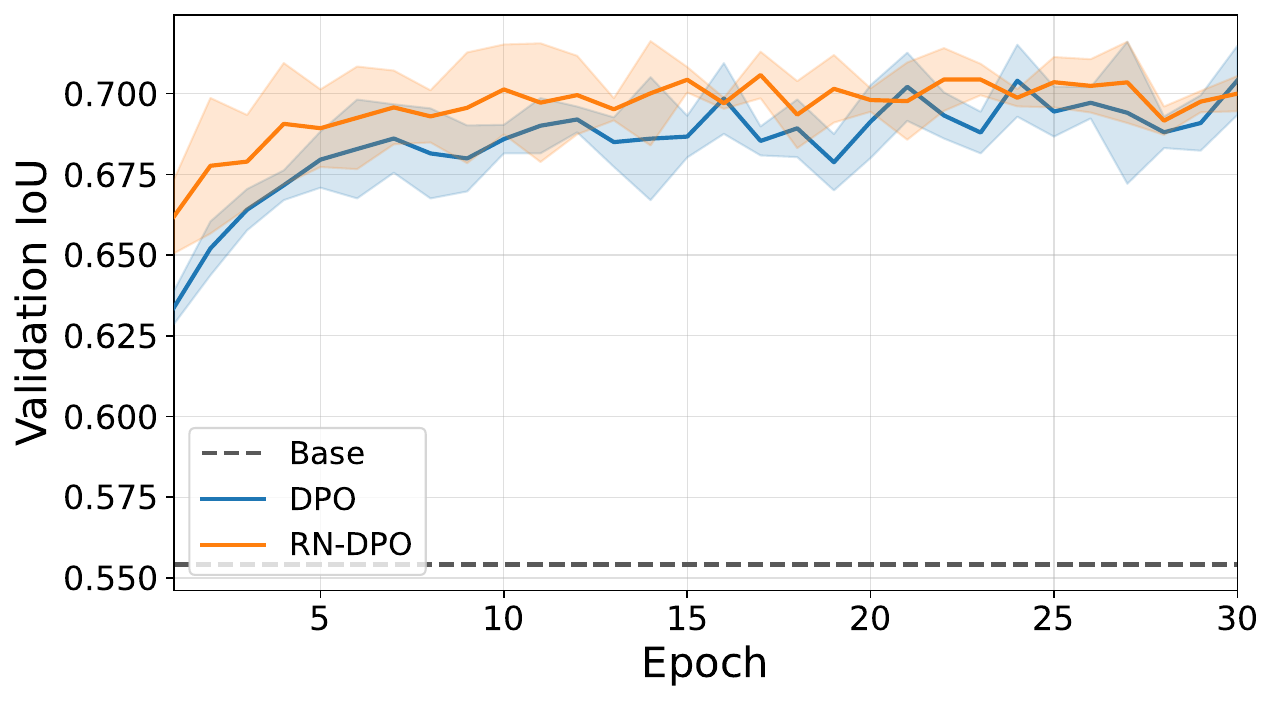}
    \subcaption*{Strong judge}
  \end{minipage}

  \caption{\textbf{ACDC}}
  \label{fig:curves_acdc}
\end{subfigure}
\vspace{1em}

\caption{\textbf{Validation IoU curves.} JSRT (top) and ACDC (bottom) shown with weak/strong judge regimes for the weak base model.}
\label{fig:val_curves_col}
\end{figure}

\subsection{Main results}\label{subsec:main}

\textbf{Training stability under noisy judges.}
Figure \ref{fig:val_curves_col} shows validation IoU trajectories for vanilla DPO and RN-DPO under weak and strong judges on both JSRT and ACDC. While vanilla DPO often achieves early gains in the weak judge regime, its performance degrades over continued fine-tuning, exhibiting monotonic drift. In contrast, RN-DPO substantially stabilizes optimization: validation IoU remains largely flat after the initial improvement phase, with only transient fluctuations rather than sustained downward trends. Under strong judges, both DPO and RN-DPO remain stable throughout fine-tuning, with RN-DPO consistenly achieving higher performance. These dynamics mirror the TailAvg improvements in Tables~\ref{tab:jsrt_main}--\ref{tab:acdc_main} and support RN-DPO as a robustness intervention that prevents the accumulation of harmful preference updates over time.

\begin{table*}[!b]
\centering
\caption{\textbf{ACDC main results} (mean$\pm$std over 3 label-draw seeds).
IoU and TailAvg are reported for various methods across four different regimes, that are characterized by weak/strong base segmenters and judges. $N_{\text{QC}}$ are chosen separately for weak and strong bases to maintain comparable judge reliability across regimes. In this case, $N_{\text{QC}}$ is referring to slices, while $N_{seg}$ defines the number of patients.}
\label{tab:acdc_main}
\setlength{\tabcolsep}{5pt}
\renewcommand{\arraystretch}{1.12}
\small


\begin{tabularx}{\textwidth}{Y l c c cc cc}
\toprule
\multirow{2}{*}{\textbf{Regime}} & \multirow{2}{*}{\textbf{Method}} &
\multirow{2}{*}{\makecell{\textbf{$\boldsymbol{N_{\text{QC}}}$}\\[-1pt] (weak/strong)}} &
\multicolumn{2}{c}{\textbf{Weak Base} ($N_{\text{seg}}{=}3$)} &
\multicolumn{2}{c}{\textbf{Strong Base} ($N_{\text{seg}}{=}5$)} \\
\cmidrule(lr){4-5}\cmidrule(lr){6-7}
& & &
\textbf{IoU} $\uparrow$ & \textbf{TailAvg} $\uparrow$ &
\textbf{IoU} $\uparrow$ & \textbf{TailAvg} $\uparrow$ \\
\midrule

\multicolumn{2}{l}{\textbf{Supervised baseline}} &
-- &
0.573$\pm$0.056 & -- &
0.676$\pm$0.020 & -- \\
\midrule

\multirow{7}{*}{\textbf{Weak judge}}
& Select-best         & \multirow{7}{*}{\makecell{32/46}} &
0.599$\pm$0.047  & -- &
0.709$\pm$0.16 & -- \\
& Top-1 Pseudolabeling    & &
0.619$\pm$0.02 & 0.514$\pm$0.019  &
0.699$\pm$0.002 & 0.653$\pm$0.016  \\
& DPO \cite{rafailov2023direct}                  & &
0.601$\pm$0.014 & 0.482$\pm$0.018 &
0.701$\pm$0.006 & 0.645$\pm$0.009 \\
& DPO (rand.)         & &
0.624$\pm$0.024 & 0.495$\pm$0.004 &
0.707$\pm$0.004 & 0.568$\pm$0.010 \\
& IPO \cite{azar2024general}      & &
0.605$\pm$0.021 & 0.517$\pm$0.017 &
0.702$\pm$0.005 & 0.646$\pm$0.010 \\
& rDPO \cite{chowdhury2024provably}                & &
0.596$\pm$0.015 & 0.491$\pm$0.015 &
0.699$\pm$0.008 & 0.645$\pm$0.014 \\
& RN-DPO              & &
0.633$\pm$0.012 & 0.581$\pm$0.023 &
\textbf{0.727$\pm$0.013} & \textbf{0.708$\pm$0.017} \\
& RN-DPO (rand.)      & &
\textbf{0.637$\pm$0.036} & \textbf{0.592$\pm$0.017} &
0.722$\pm$0.013 & 0.702$\pm$0.015 \\
\midrule

\multirow{7}{*}{\textbf{Strong judge}}
& Select-best         & \multirow{7}{*}{\makecell{66/124}} &
0.624$\pm$0.048 & -- &
0.722$\pm$0.019  & -- \\
& Top-1 Pseudolabeling          & &
0.708$\pm$0.009 & 0.700$\pm$0.013 &
0.748$\pm$0.012 & 0.726$\pm$0.010 \\
& DPO \cite{rafailov2023direct}                & &
0.716$\pm$0.003 & 0.695$\pm$0.006 &
0.741$\pm$0.008 & 0.720$\pm$0.010 \\
& DPO (rand.)         & &
0.704$\pm$0.019 & 0.595$\pm$0.034 &
0.732$\pm$0.010 & 0.619$\pm$0.011 \\
& IPO \cite{azar2024general}      & &
0.715$\pm$0.010 & 0.685$\pm$0.005 &
0.750$\pm$0.012 & 0.704$\pm$0.016 \\
& rDPO \cite{chowdhury2024provably}               & &
\textbf{0.718$\pm$0.005} & 0.696$\pm$0.001 &
0.746$\pm$0.011 & 0.719$\pm$0.019 \\
& RN-DPO              & &
0.716$\pm$0.005 & \textbf{0.700$\pm$0.005} &
\textbf{0.753$\pm$0.015} & 0.739$\pm$0.015 \\
& RN-DPO (rand.)      & &
0.717$\pm$0.013 & 0.699$\pm$0.004 &
0.751$\pm$0.007 & \textbf{0.741$\pm$0.012} \\
\midrule

\multirow{2}{*}{\textbf{GT oracle}}
& DPO                 & -- &
0.728$\pm$0.010 & 0.731$\pm$0.005 &
0.748$\pm$0.006 & 0.748$\pm$0.003 \\
& RN-DPO              & -- &
\textbf{0.753$\pm$0.009} & \textbf{0.748$\pm$0.009} &
\textbf{0.773$\pm$0.013} & \textbf{0.767$\pm$0.007} \\
\bottomrule
\end{tabularx}
\end{table*} 

\begin{figure}[t]
    \centering
    \includegraphics[width=\columnwidth]{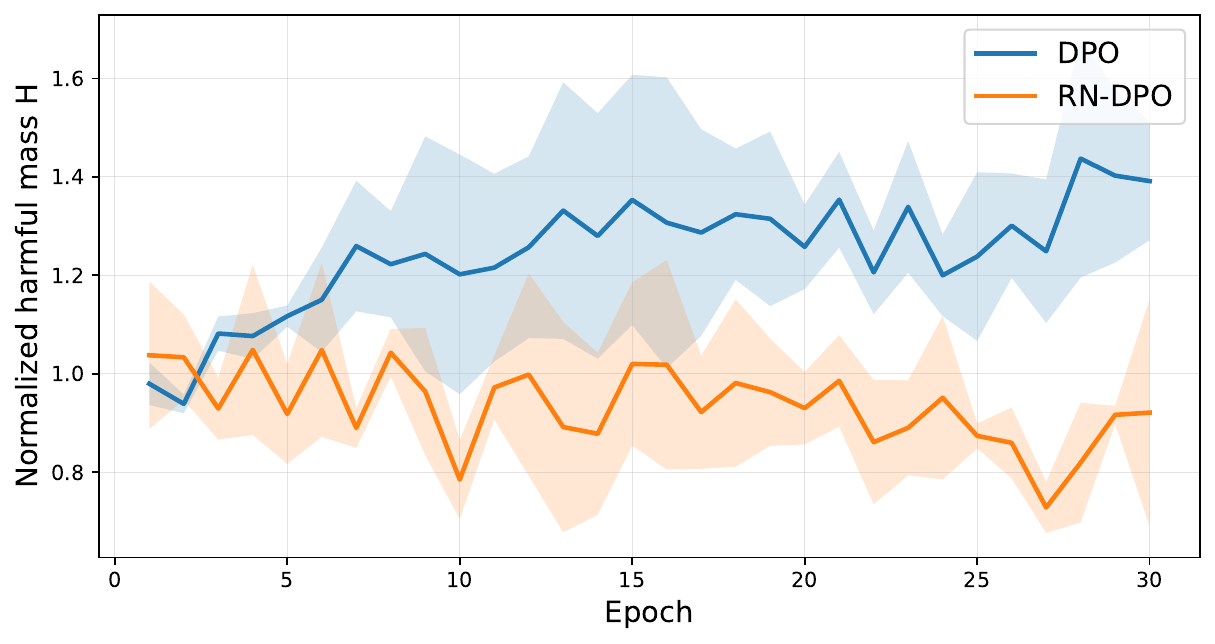}
    \caption{\textbf{Harmful update mass over training (JSRT, weak base/weak judge).}
    We track the normalized harmful update mass $H$ (Eq.~\ref{eq:harm_mass}), which aggregates oracle-negative preference margins weighted by disagreement fraction and DPO update strength.}
    \label{fig:harmful_mass}
\end{figure}

\textbf{Harmful update mass and training stability.}
To connect judge noise to training stability, we track a per-epoch \emph{harmful update mass} proxy measuring how much erroneous preference signal is applied during fine-tuning.
For each mined pair $(y^+,y^-)$ we compute an oracle margin $\Delta^* = m(y^+) - m(y^-)$, where $m(\cdot)$ denotes IoU w.r.t.\ ground truth (used for analysis only).
Pairs with $\Delta^*<0$ correspond to harmful supervision.
We aggregate harmful mass as
\begin{equation}
\label{eq:harm_mass}
H=\mathbb{E}\big[\mathbbm{1}[\Delta^*<0]\cdot(-\Delta^*)\cdot w\big],
\end{equation}
where $w=\beta\,\sigma(-\beta\Delta)$ is the DPO per-pair update-strength factor, with $\Delta$ the DPO log-ratio difference for the pair. 
Figure~\ref{fig:harmful_mass} plots $H$ over preference fine-tuning epochs, computed on the pairs mined during training time and normalized by the mean of the first three epochs.
Vanilla DPO shows a systematic increase in harmful mass over time, consistent with accumulating harmful supervision, whereas RN-DPO keeps harmful mass bounded and often drives it downward, matching its improved stability.
\setlength{\abovecaptionskip}{0pt}
\setlength{\belowcaptionskip}{-10pt}

\setlength{\abovecaptionskip}{2pt}
\setlength{\belowcaptionskip}{0pt}

\begin{table}[t]
\centering
\caption{\textbf{Proposal budget ablation.} Performance for noisy base and judge setting for varying number $K$ of generated proposals on JSRT. Mean$\pm$std over 3 label-draw seeds.}
\label{tab:k_ablation}
\setlength{\tabcolsep}{8pt}
\renewcommand{\arraystretch}{1.15}

\begin{tabularx}{\linewidth}{c c >{\centering\arraybackslash}X >{\centering\arraybackslash}X}
\toprule
\textbf{$K$} & \textbf{Method} & \textbf{IoU $\uparrow$} & \textbf{TailAvg $\uparrow$} \\
\midrule

\multirow{2}{*}{8}
 & DPO     & \makecell{0.607$\pm$0.004} & \makecell{0.594$\pm$0.011} \\
 & RN-DPO  & \makecell{\textbf{0.638}$\boldsymbol{\pm}$\textbf{0.010}} & \makecell{\textbf{0.621}$\boldsymbol{\pm}$\textbf{0.007}} \\
\midrule

\multirow{2}{*}{16}
 & DPO     & \makecell{0.606$\pm$0.007} & \makecell{0.593$\pm$0.005} \\
 & RN-DPO  & \makecell{\textbf{0.637}$\boldsymbol{\pm}$\textbf{0.014}} & \makecell{\textbf{0.627}$\boldsymbol{\pm}$\textbf{0.010}} \\
\midrule

\multirow{2}{*}{32}
 & DPO     & \makecell{0.604$\pm$0.002} & \makecell{0.565$\pm$0.011} \\
 & RN-DPO  & \makecell{\textbf{0.631}$\boldsymbol{\pm}$\textbf{0.012}} & \makecell{\textbf{0.619}$\boldsymbol{\pm}$\textbf{0.009}} \\
\bottomrule

\end{tabularx}
\end{table} 
\begin{table}[b]
\centering
\caption{\textbf{Judge ablations on JSRT (weak base, weak judge)}. Entries are means over 3 label-draw seeds.}
\label{tab:judge_ablations_jsrt_weak}
\setlength{\tabcolsep}{6pt}
\renewcommand{\arraystretch}{1.08}
\small
\begin{tabular}{llcc}
\toprule
\textbf{Judge choice} & \textbf{Method} & \textbf{IoU} $\uparrow$ & \textbf{TailAvg} $\uparrow$ \\
\midrule
-- & Supervised & 0.558$\pm$0.017 & / \\
\midrule
\multirow{2}{*}{Ensemble} & DPO & 0.615$\pm$0.013 & 0.591$\pm$0.013 \\
& RN-DPO & \textbf{0.655$\pm$0.002} & \textbf{0.644$\pm$0.029} \\
\midrule
\multirow{2}{*}{IoU regressor} & DPO & 0.615$\pm$0.007 & 0.585$\pm$0.017 \\
& RN-DPO & \textbf{0.656$\pm$0.019} & \textbf{0.644$\pm$0.020} \\
\bottomrule
\end{tabular}
\end{table}

\setlength{\abovecaptionskip}{0pt}
\setlength{\belowcaptionskip}{-10pt}
\begin{figure*}[t]
    \centering
    \includegraphics[width=0.95\textwidth]{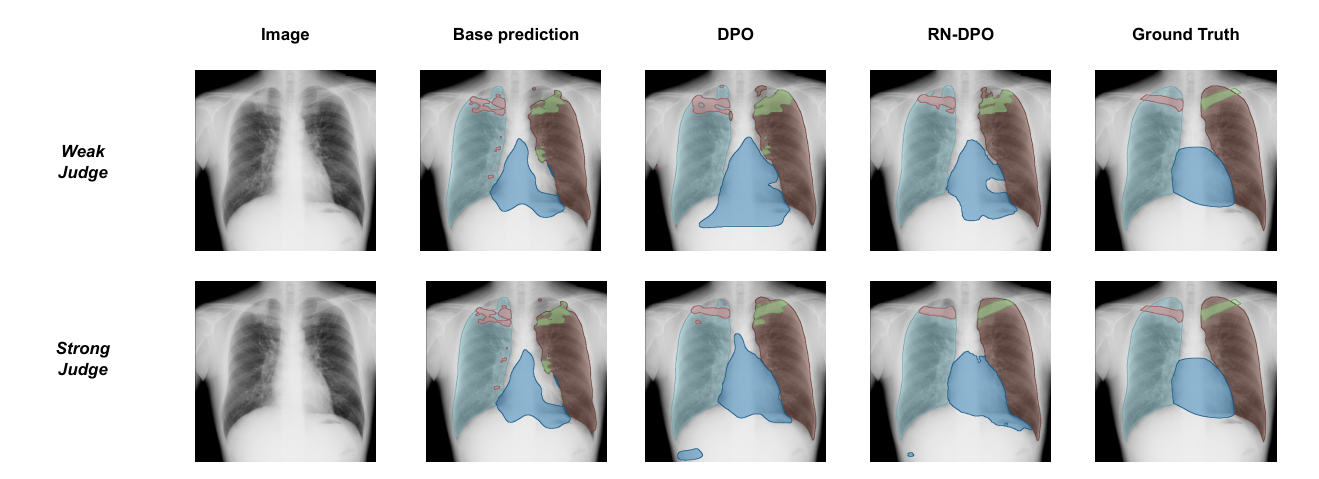}
\caption{\textbf{Qualitative segmentation results on JSRT.}
Examples comparing the output segmentations from the base segmenter with models fine-tuned using DPO and RN-DPO for the weak base regime.}
    \label{fig:qual}
\end{figure*}

\textbf{Comparison against related work.}
Table \ref{tab:jsrt_main} and Table \ref{tab:acdc_main} describe the results for preference fine-tuning performance under controlled base and judge regimes for both JSRT and ACDC respectively. 
Across all regimes for both datasets, RN-DPO consistently improves both peak accuracy and long-horizon stability compared to vanilla DPO.  Under weak judges, where noisy rankings induce drift, RN-DPO substantially improves TailAvg, e.g. from $0.565$ to $0.619$ on the weak base for JSRT and $0.482$ to $0.581$ for ACDC. For both weak and strong bases, it also shows significant performance boosts for both datasets. Under strong judges, RN-DPO remains beneficial, especially on JSRT, where it improves TailAvg from $0.679$ to $0.704$ on the weak base (+0.025) and from $0.743$ to $0.784$ on the strong base (+0.041).
With strong judges, random mining becomes less competitive, aligning with the intuition that headroom-seeking top-based mining is safer when the judge is reliable.
Notably, rDPO provides little or no improvement over DPO in this setting, highlighting that segmentation-specific robustness challenges are not fully addressed by existing preference-noise corrections.  
Pseudolabeling is a surprisingly strong baseline, especially under strong judges, occasionally matching or exceeding DPO, while RN-DPO matches or surpasses it in most regimes.
Finally, in the GT oracle setting, where we assume a perfect judge, RN-DPO continues to deliver large gains over DPO  suggesting that region normalization improves optimization even when preferences are perfectly correct, by better exploiting fine-grained differences. We report on the per-class performance in the Appendix~\ref{apx:performance}.
For the same regimes, Fig. \ref{fig:qual} provides qualitative comparisons between the base prediction and models fine-tuned with DPO and RN-DPO. RN-DPO yields the most visible improvements on the more challenging structures (heart and clavicles), correcting fragmented regions and reducing spurious artifacts.

\subsection{Ablations}\label{subsec:ablations}

\paragraph{Number of proposals.}
Table~\ref{tab:k_ablation} ablates the proposal budget $K$ in the noisy base/judge regime.
RN-DPO consistently outperforms vanilla DPO across all tested slate sizes ($K\in\{8,16,32\}$), improving both Peak IoU and TailAvg in every setting.
Importantly, the gains remain stable as $K$ varies, indicating that RN-DPO does not rely on a specific proposal budget to be effective.

\paragraph{Judge ablation.}

Ensemble agreement is a widely used proxy for segmentation reliability and a practical mechanism for automated quality control
\cite{zenk2025comparative,hann2021deep}. To instantiate a model-based QC signal \emph{without} introducing an additional QC-labeled split,
we construct an \emph{ensemble-agreement judge} by training $M{=}5$ segmenters with different random seeds on the same labeled set used for Stage~1.
Given an image $x$ and a proposal $y^{(k)}$, we score the proposal by its mean agreement to the ensemble predictions,
\begin{equation}
s^{(k)} \;=\; \frac{1}{M}\sum_{m=1}^{M}\mathrm{IoU}\!\left(y^{(k)}, \hat{y}^{(m)}(x)\right),
\end{equation}
and rank the slate using $s^{(k)}$ to mine preference pairs.
As a complementary ablation, we replace mask-to-mask agreement with an \emph{IoU regressor} that predicts proposal quality from $(x,y^{(k)})$,
providing a qualitatively different judge interface than ensemble overlap.
Table~4 shows that RN-DPO improves both peak performance and stability under \emph{both} alternative QC signals, suggesting that RN-DPO yields robustness beyond the single-judge setting.

\section{Conclusion}\label{sec:results}

We conducted an extensive study of preference-based fine-tuning for medical image segmentation under noisy  judges. Our results show that DPO performance depends on the the mining strategy, and the judge reliability, and that performance drift can arise from high-leverage mis-rankings. To address this, we introduced Region-Normalized DPO (RN-DPO), a segmentation-aware objective that normalizes preference updates over the disagreement region between candidate masks, reducing the impact of harmful comparisons. Across two datasets and multiple base/judge regimes, RN-DPO consistently improves both peak IoU and sustained performance over standard DPO and outperforms strong baselines across most regimes. Overall, our results highlight disagreement-region normalization as a broadly applicable design principle for preference-based learning in medical image segmentation.

\section{Impact Statement}
This work reduces reliance on costly pixel-wise annotations by enabling stable preference-based fine-tuning for medical image segmentation using noisy automatic quality-control signals. By introducing Region-Normalized DPO, we mitigate the risk of harmful updates that can arise when such signals are biased or unreliable, improving robustness in a high-stakes medical setting.

The proposed approach can lower barriers to developing medical imaging systems, supporting broader adoption in data-scarce and resource-limited environments. At the same time, our findings highlight the importance of guarding against error amplification and hidden biases when learning from automated judges, reinforcing the need for careful validation and continued human oversight in clinical deployment.

\nocite{langley00}

\bibliography{literature}
\bibliographystyle{icml2026}

\newpage
\appendix
\onecolumn
\section{Appendix}

\subsection{Implementation Details and Hyperparameters}
\label{app:hpms}

\setlength{\abovecaptionskip}{0pt}
\setlength{\belowcaptionskip}{4pt}
\begin{table}[t]
\centering
\small
\caption{\textbf{JSRT hyperparameter sweep.} Each entry reports the selection score
$\tfrac{1}{2}(\text{Peak IoU}+\text{TailAvg})$ on the validation set; we choose the best setting (bold) per method and judge regime.}
\setlength{\tabcolsep}{4pt}
\begin{tabular}{c c | ccc | ccc}
\toprule
\multirow{2}{*}{Judge} & \multirow{2}{*}{$\beta$}
& \multicolumn{3}{c|}{Vanilla DPO (Score)}
& \multicolumn{3}{c}{RN-DPO (Score)} \\
& & $2\!\times\!10^{-5}$ & $5\!\times\!10^{-5}$ & $10^{-4}$
  & $2\!\times\!10^{-5}$ & $5\!\times\!10^{-5}$ & $10^{-4}$ \\
\midrule

\multirow{3}{*}{Weak}  & 0.25 & 0.576 & 0.552 & 0.526 & 0.609 & 0.596 & 0.574 \\
                      & 1.00 & 0.572 & 0.561 & 0.562 & 0.635 & 0.643 & 0.639  \\
                      & 1.50 & \textbf{0.577} & 0.567 & 0.556 & 0.639 & \textbf{0.643} & 0.642  \\
\midrule
\multirow{3}{*}{Strong}& 0.25 & 0.676  & 0.678 & 0.675 & 0.719 & 0.718 & \textbf{0.723} \\
                       & 1.00 & 0.677 & 0.683 & 0.679 & 0.701 & 0.714 & 0.720 \\
                       & 1.50 & 0.681 & \textbf{0.692} & 0.680 & 0.695 & 0.709 & 0.719 \\
\bottomrule
\end{tabular}

\label{tab:hparam_jsrt}
\end{table}
\begin{table}[t]
\centering
\small
\setlength{\tabcolsep}{4pt}

\caption{\textbf{ACDC hyperparameter sweep.} Each entry reports the selection score
$\tfrac{1}{2}(\text{Peak IoU}+\text{TailAvg})$ on the validation set; we choose the best setting (bold) per method and judge regime.}
\begin{tabular}{c c | ccc | ccc}
\toprule

\multirow{2}{*}{Judge} & \multirow{2}{*}{$\beta$}
& \multicolumn{3}{c|}{Vanilla DPO (Score)}
& \multicolumn{3}{c}{RN-DPO (Score)} \\
& & $2\!\times\!10^{-5}$ & $5\!\times\!10^{-5}$ & $10^{-4}$
  & $2\!\times\!10^{-5}$ & $5\!\times\!10^{-5}$ & $10^{-4}$ \\
\midrule

\multirow{3}{*}{Weak}  & 0.25 & 0.519 & 0.507 & 0.521 & 0.564 & 0.549 & 0.527 \\
                      & 1.00 & 0.520 & 0.512 & 0.508 & 0.583 & 0.579 & 0.583  \\
                      & 1.50 & \textbf{0.547} & 0.518 & 0.556 & 0.585 & 0.590 & \textbf{0.595}  \\
\midrule
\multirow{3}{*}{Strong}& 0.25 & \textbf{0.705}  & 0.699 & 0.685 & 0.697 & \textbf{0.707} & 0.698 \\
                       & 1.00 & 0.700 & 0.699 & 0.684 & 0.664& 0.673 & 0.690 \\
                       & 1.50 & 0.704 & 0.695 & 0.684 & 0.657 & 0.672 & 0.679 \\
\bottomrule
\end{tabular}
\label{tab:hparam_acdc}
\end{table}

All experiments were run on NVIDIA A100 80GB GPUs. Each training run used a single GPU and takes approximately 1 hour. 
We perform an extensive hyperparameter grid on both datasets, separately tuning standard DPO and RN-DPO under weak and strong judges, as these settings can prefer slightly different learning rate and $\beta$ choices. Table \ref{tab:hparam_jsrt} and \ref{tab:hparam_acdc} show the full grid search and selected configurations.
Additionally, we evaluate performance using a cosine schedule (instead of a constant schedule) in Table \ref{tab:schedule_adamw_ablation}.

\begin{table}[!t]
\centering
\small

\caption{Effect of learning-rate scheduling choice on vanilla DPO vs.\ RN-DPO.}

\begin{tabular}{lccc ccc}
\toprule
& \multicolumn{3}{c}{Vanilla DPO} & \multicolumn{3}{c}{RN-DPO} \\
\cmidrule(lr){2-4}\cmidrule(lr){5-7}
Stage-2 setting & TailAvg & IoU & Avg & TailAvg & IoU & Avg \\
\midrule
Standard (constant LR) & 0.550 & 0.607 & 0.578 & 0.636 & 0.651 & 0.643 \\
Cosine schedule & 0.570 & 0.602 & 0.586 & 0.632 & 0.647 & 0.640 \\
\bottomrule
\end{tabular}
\label{tab:schedule_adamw_ablation}
\end{table}

\subsection{Proposal slate generation}
\label{app:proposals}

For each image, we construct a \emph{slate} of candidate segmentations intended to provide
(i) \textbf{headroom} (some candidates plausibly better than the current prediction) and
(ii) \textbf{structured diversity} (meaningful variations in topology and boundary placement).
The same proposal families are used for both datasets, with minor adaptations for \textbf{JSRT} (multi-label) versus \textbf{ACDC} (multi-class).

\smallskip
\noindent\textbf{Representation.}
For \textbf{JSRT}, proposals are generated per class as binary masks and stacked across classes.
For \textbf{ACDC}, proposals are generated as single-label maps, and all operations preserve mutual exclusivity between classes.

\smallskip
\noindent\textbf{(0) Base prediction.}
We always include the model's unmodified prediction as an anchor proposal.

\smallskip
\noindent\textbf{(1) Topology edits (components, holes).}
Starting from the anchor, we generate topology-aware variants:
(i) \emph{hole filling} (fill small holes, with size relative to the predicted structure and capped to avoid overly aggressive filling), and
(ii) \emph{component cleanup + hole filling} (keep only the largest connected component per class, then fill holes).
For \textbf{JSRT}, these are applied independently per class.
For \textbf{ACDC}, they are applied per foreground class on the label map; importantly, hole filling only assigns pixels that are currently
background and does not overwrite other foreground classes.

\smallskip
\noindent\textbf{(2) Monte Carlo dropout.}
We enable dropout at inference time in the final decoder module and segmentation head, run multiple stochastic forward passes, and include both
individual stochastic predictions and their mean as proposals.

\smallskip
\noindent\textbf{(3) Test-time augmentation (TTA).}
We run the model under lightweight intensity augmentations (e.g., gamma, brightness, contrast) and include only the mean prediction across
augmentations.

\smallskip
\noindent\textbf{(4) Controlled extent / class-bias variants.}
We include proposals that induce controlled under-/over-segmentation without introducing arbitrary shapes.
For \textbf{JSRT}, we use \emph{area-quantile thresholding}: adjust the binarization threshold so the foreground area matches a small set of
multiplicative scalings of the anchor area (per class), preserving confidence ordering while changing extent.
For \textbf{ACDC}, we use \emph{logit-bias perturbations}: shift logits to change the relative preference between foreground and background, or a
single foreground class versus the remaining classes, yielding controlled changes near ambiguous boundaries.

\smallskip
\noindent\textbf{(5) Signed-distance boundary offsets.}
We generate explicit boundary-shift proposals by applying small signed-distance offsets, producing systematic shrink/expand variants. In multi-class
settings, this can be applied uniformly per class or targeted to salient interfaces (chosen by domain knowledge or simple heuristics such as frequent
class confusion).
For \textbf{JSRT}, we apply per-class inward/outward offsets.
For \textbf{ACDC}, we additionally apply interface-focused offsets for commonly ambiguous boundaries (e.g., LV--MYO and MYO--background) and include
RV-focused shrink/expand variants, using conservative conflict resolution so expansions do not overwrite unrelated structures.

\smallskip
\noindent\textbf{(6) Compositions.}
To increase headroom, we include a small number of composed transforms, such as:
topology $\rightarrow$ boundary shift (boundary offsets applied after topology cleanup), MC-mean $\rightarrow$ topology (cleanup applied to the MC-dropout
mean), and optionally applying topology cleanup to extreme shrink/expand variants to suppress artifacts.
Example proposals for JSRT can be seen in Fig. 5.

\begin{figure*}[t]
    \centering
    \includegraphics[width=\textwidth]{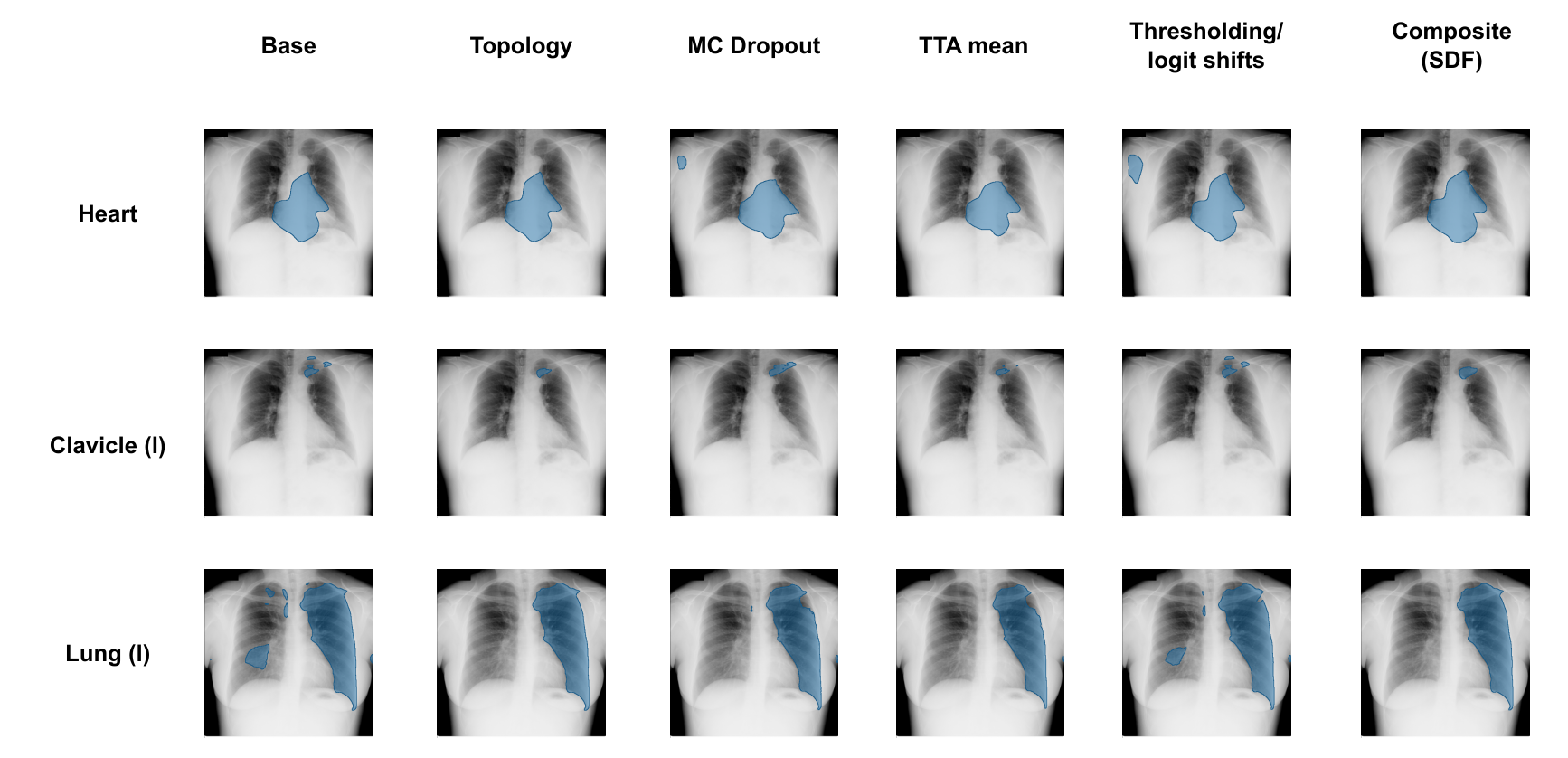}
\caption{\textbf{Qualitative segmentation proposals using different strategies on JSRT.} (SDF = signed distance fields)}
    \label{fig:qual}
\end{figure*}

\begin{table*}[!t]
\centering
\caption{\textbf{JSRT per-anatomy results} (mean$\pm$std over seeds). IoU results for DPO and RN-DPO across all five anatomies (left/right clavicles (Clav-L, Clav-R), left/right lungs (Lung-L, Lung-R) and heart. Supervised denotes the performance of the base segmenter.}
\label{tab:jsrt_per_anatomy_base}
\setlength{\tabcolsep}{3.5pt}
\renewcommand{\arraystretch}{1.05}
\scriptsize

\begin{tabular}{l l c c c c c}
\toprule
\textbf{Base} & \textbf{Method} &
\textbf{Clav-L} &
\textbf{Clav-R} &
\textbf{Lung-L} &
\textbf{Lung-R} &
\textbf{Heart} \\
\midrule

\textbf{--}
& Supervised & 0.355$\pm$0.055 & 0.389$\pm$0.038 & 0.756$\pm$0.027 & 0.808$\pm$0.031 & 0.482$\pm$0.110 \\
\midrule

\multirow{2}{*}{\textbf{Weak}}
& DPO        & 0.376$\pm$0.057 & 0.444$\pm$0.063 & 0.836$\pm$0.018 & 0.879$\pm$0.032 & 0.483$\pm$0.050 \\
& RN-DPO     & \textbf{0.433}$\boldsymbol{\pm}$\textbf{0.047} &
               \textbf{0.463}$\boldsymbol{\pm}$\textbf{0.039} &
               \textbf{0.844}$\boldsymbol{\pm}$\textbf{0.011} &
               \textbf{0.896}$\boldsymbol{\pm}$\textbf{0.018} &
               \textbf{0.520}$\boldsymbol{\pm}$\textbf{0.092} \\
\midrule

\multirow{2}{*}{\textbf{Strong}}
& DPO        & 0.500$\pm$0.023 & 0.543$\pm$0.039 & 0.869$\pm$0.006 & 0.898$\pm$0.001 & 0.668$\pm$0.024 \\
& RN-DPO     & \textbf{0.557}$\boldsymbol{\pm}$\textbf{0.008} &
               \textbf{0.584}$\boldsymbol{\pm}$\textbf{0.003} &
               \textbf{0.882}$\boldsymbol{\pm}$\textbf{0.011} &
               \textbf{0.906}$\boldsymbol{\pm}$\textbf{0.010} &
               \textbf{0.675}$\boldsymbol{\pm}$\textbf{0.020} \\
\bottomrule
\end{tabular}
\end{table*}
\begin{table*}[!t]
\centering
\caption{\textbf{ACDC per-class results} (mean$\pm$std over seeds). IoU results for DPO and RN-DPO across all cardiac structures: right ventricle (RV), myocardium (Myo), and left ventricle (LV). "Supervised" denotes the performance of the base segmenter.}
\label{tab:acdc_per_anatomy_base}
\setlength{\tabcolsep}{3.5pt}
\renewcommand{\arraystretch}{1.05}
\scriptsize

\begin{tabular}{l l c c c}
\toprule
\textbf{Base} & \textbf{Method} &
\textbf{RV} &
\textbf{Myo} &
\textbf{LV} \\
\midrule

\textbf{--}
& Supervised & 0.524$\pm$0.081 & 0.516$\pm$0.041 & 0.678$\pm$0.048 \\
\midrule

\multirow{2}{*}{\textbf{Weak}}
& DPO        & 0.556$\pm$0.026 & 0.548$\pm$0.018 & 0.700$\pm$0.015 \\
& RN-DPO     & \textbf{0.584}$\boldsymbol{\pm}$\textbf{0.015} &
               \textbf{0.574}$\boldsymbol{\pm}$\textbf{0.008} &
               \textbf{0.742}$\boldsymbol{\pm}$\textbf{0.015} \\
               \midrule
\multirow{2}{*}{\textbf{Strong}}
& DPO        & \textbf{0.712}$\boldsymbol{\pm}$\textbf{0.008} & 0.644$\pm$0.003 & 0.797$\pm$0.004 \\
& RN-DPO     & 0.696$\pm$0.010 &
               \textbf{0.648}$\boldsymbol{\pm}$\textbf{0.011} &
               \textbf{0.803}$\boldsymbol{\pm}$\textbf{0.009} \\
\bottomrule
\end{tabular}
\end{table*}

\subsection{Class-wise Performance}\label{apx:performance}
Table \ref{tab:jsrt_per_anatomy_base} and Table \ref{tab:acdc_per_anatomy_base} reports per-anatomy IoU on JSRT for weak and strong base segmenters under the weak-judge regime. RN-DPO improves IoU across all five structures, with the largest gains on the more challenging, low-IoU anatomies (clavicles and heart).

\subsection{Judge Diagnostics}\label{app:diagnostics}
\begin{table}[t]
\centering
\small
\caption{Judge characterization via slate-level diagnostics over the validation set.}
\setlength{\tabcolsep}{5pt}
\begin{tabular}{lccccc}
\toprule
\textbf{Regime} &
\textbf{HarmTop$\downarrow$} &
\textbf{HarmMag$\downarrow$} &
\textbf{Headroom$_{\text{judge}}$$\uparrow$} &
\textbf{Regret$\downarrow$} &
\textbf{PairFlip$\downarrow$} \\
\midrule
\multicolumn{6}{l}{\textbf{JSRT}}\\
Weak base / Weak judge   & 0.222 & 0.010  & 0.048 & 0.019 & 0.111  \\
Weak base / Strong judge & 0.069 & 0.001 &  0.060 & 0.01 & 0.069 \\
Strong base / Weak judge & 0.267 & 0.008  &  0.018 & 0.038 & 0.2  \\
Strong base / Strong judge & 0.267 & 0.005 & 0.039 & 0.017  & 0.2 \\
\midrule
\multicolumn{6}{l}{\textbf{ACDC}}\\
Weak base / Weak judge   & 0.386  & 0.017  & 0.009  & 0.035 & 0.331  \\
Weak base / Strong judge & 0.192  & 0.004  & 0.032 & 0.012 & 0.126  \\
Strong base / Weak judge & 0.378  & 0.011  & 0.020 & 0.030 & 0.295  \\
Strong base / Strong judge & 0.161 & 0.003  & 0.038  & 0.012  & 0.114  \\
\bottomrule
\end{tabular}
\label{tab:judge_metrics}
\end{table}

For each image, let $\{y^{(k)}\}_{k=1}^K$ denote the proposal slate, let $y^{\mathrm{base}}$ be the anchor (model) proposal, and let $y^{\mathrm{top}}$ be the judge-selected top proposal.
Let $m(\cdot)$ denote oracle mask quality measured by IoU w.r.t.\ ground truth (used only for post-hoc analysis).

\begin{itemize}
\item \textbf{HarmTop$\downarrow$ (harmful-top rate).}
\[
\mathrm{HarmTop} \;=\; \Pr\!\left[m\!\left(y^{\mathrm{top}}\right) < m\!\left(y^{\mathrm{base}}\right)\right].
\]

\item \textbf{HarmMag$\downarrow$ (harmful-top magnitude).}
\[
\mathrm{HarmMag} \;=\; \mathbb{E}\!\left[\big(m(y^{\mathrm{base}})-m(y^{\mathrm{top}})\big)_{+}\right],
\qquad (a)_{+}=\max(a,0).
\]

\item \textbf{Headroom$_{\text{judge}}$$\uparrow$ (achieved headroom).}
Let $y^{\mathrm{base}}$ denote the base (anchor) proposal and $y^{\mathrm{top}}$ the judge-selected top proposal in the slate. We define the achieved headroom as
\[
\mathrm{Headroom}_{\text{judge}} \;=\; m\!\left(y^{\mathrm{top}}\right) - m\!\left(y^{\mathrm{base}}\right),
\]
where $m(\cdot)$ denotes oracle IoU w.r.t.\ ground truth (used for analysis only).

\item \textbf{Regret$\downarrow$ (oracle regret).}
\[
\mathrm{Regret} \;=\; m(y^{*}) - m(y^{\mathrm{top}}).
\]

\item \textbf{PairFlip$\downarrow$ (miner-aligned flip rate).}
Let $(y^{+},y^{-})$ be the mined training pair (e.g., top-vs-random under the judge ranking). Define the oracle pair margin
$\Delta^*_{\mathrm{pair}} = m(y^{+})-m(y^{-})$.
Then
\[
\mathrm{PairFlip} \;=\; \Pr\!\left[\Delta^*_{\mathrm{pair}}<0\right].
\]
\end{itemize}

\noindent We compute each metric per slate using the validation set. 



\end{document}